\documentclass{article}
\pdfoutput=1

\usepackage{arxiv}
\usepackage[export]{adjustbox}
\usepackage[para,online,flushleft]{threeparttable}
\usepackage{booktabs}
\usepackage{threeparttable}
\usepackage{makecell}
\usepackage[utf8]{inputenc} 
\usepackage[T1]{fontenc}    
\usepackage{hyperref}       
\usepackage{url}            
\usepackage{booktabs}       
\usepackage{amsfonts}       
\usepackage{nicefrac}       
\usepackage{microtype}      
\usepackage{graphicx}
\usepackage[caption=false]{subfig}
\usepackage{doi}
\usepackage{amsmath}
\usepackage{multirow}
\usepackage{multicol}
\usepackage{titlesec}
\titlespacing*{\section}{0pt}{0.3\baselineskip}{0.3\baselineskip}
\titlespacing*{\paragraph}{0pt}{0.3\baselineskip}{0.3\baselineskip}

\title{A Study on Reproducibility and Replicability of Table Structure Recognition Methods}

\author{Kehinde Ajayi\\
	Old Dominion University\\
	Norfolk, VA 23529 \\
	\texttt{kajay001@odu.edu} \\
    \And
    Muntabir Hasan Choudhury\\
    Old Dominion University\\
    Norfolk, VA 23529 \\
    \texttt{mchou001@odu.edu} \\
   \And
    Sarah M. Rajtmajer\\
    IST, Pennsylvania State University\\
    University Park PA, USA\\
    \texttt{smr48@pdu.edu} \\
	\And
    Jian Wu\\
	Old Dominion University\\
	Norfolk, VA 23529 \\
	\texttt{j1wu@odu.edu} \\
}

\renewcommand{\shorttitle}{A Study on Reproducibility and Replicability of Table Structure Recognition Methods}

\begin{document}
\maketitle

\begin{abstract}
Concerns about reproducibility in artificial intelligence (AI) have emerged, as researchers have reported unsuccessful attempts to directly reproduce published findings in the field. Replicability, the ability to affirm a finding using the same procedures on new data, has not been well studied. In this paper, we examine both reproducibility and replicability of a corpus of 16 papers on table structure recognition (TSR), an AI task aimed at identifying cell locations of tables in digital documents. We attempt to reproduce published results using codes and datasets provided by the original authors. We then examine replicability using a dataset similar to the original as well as a new dataset, GenTSR, consisting of 386 annotated tables extracted from scientific papers. Out of 16 papers studied, we reproduce results consistent with the original in only four. Two of the four papers are identified as replicable using the similar dataset under certain IoU values. No paper is identified as replicable using the new dataset. We offer observations on the causes of irreproducibility and irreplicability. All code and data are available on Codeocean at \url{https://codeocean.com/capsule/6680116/tree}.
\end{abstract}


\section{Introduction}
Concerns about reproducibility, replicability, and generalizability (RR\&G) 
of findings in the social and behavioral sciences are now well-established 
\cite{fanelli2017pnas,baker20161500,camerer2018evaluating}. More recently, RR\&G concerns have been raised in the field of artificial intelligence (AI), e.g., \cite{olorisade2017reproducibility,raff2019step}. 
There has been inconsistent use of these terms across the literature. Here, we adopt definitions from Goodman et al. \cite{goodman2016}. 
By \emph{reproducibility}, we refer to computational repeatability -- obtaining consistent computational results using the same data, methods, code, and conditions of analysis.  While, \emph{replicability} is obtaining consistent results on a different but similar dataset using the same methods \cite{national2019reproducibility,nosek2021replicability,pineau2021improving,goodman2016}.  \emph{Generalizability} refers to obtaining consistent results in settings outside of the experimental framework \cite{goodman2016}. Each concept sets an incrementally higher standard than the previous one with reproducibility being the most basic requirement of fundamental science.

Existing studies of AI reproducibility have focused on empirical and computational AI, in which datasets, codes, and environments are essential conditions for reproduction. 
Some papers have examined the availability of certain information assumed critical to reproducibility. For example, Gunderson et al.\cite{gundersen2018state} studied reproducibility of AI research by investigating whether research papers include adequate metadata, i.e., detailed documentation of methodology. Others have investigated the availability of open-access datasets and software \cite{salsabil2022scik} and the executability of source codes \cite{pimentel2019jupyter}. Directly reproducing results provides the most convincing evidence of reproducibility but usually requires more time, effort, and domain knowledge. Raff \cite{raff2019step} conducted direct reproduction of AI papers. However, little effort has been put into the  \emph{replicability} of AI papers. 

\begin{figure}[h!]
    \centering
    \includegraphics[scale = 0.62]{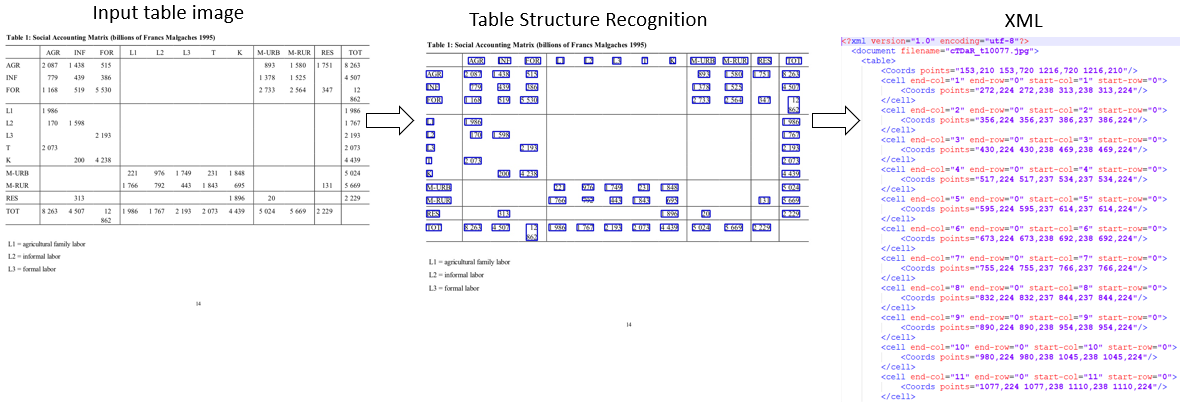}
    \caption{An illustration of the TSR problem.}\label{tsr}
\end{figure}

In this work, we investigate the reproducibility and replicability of methods for table structure recognition, an AI task aimed at parsing tables in digital documents and automatically identifying rows, columns, and cell positions in a detected table image within a document \cite{cascadetabnet}. This task is different from a related task called table detection, automatically locating tables in document images \cite{gatos2005automatic}. Earlier methods attempted to solve these two tasks separately \cite{DeepDsrt}. Recently, several end-to-end solutions based on neural networks have been proposed \cite{cascadetabnet,Multi-Type-TD-TSR,Graph-based-TSR-lee}. The input of the TSR task is a table image and the output is usually an XML or JSON file containing coordinates of detected cells (row and column numbers and pixels of cell bounding boxes). The content of cells is not identified. Figure~\ref{tsr} illustrates the TSR problem. To the best of our knowledge, our work takes the first step to assess the replicability (based on the definition given above) of published research in the domain of document analysis and pattern recognition.

The goal of our work is twofold. First, we test \emph{reproducibility} of published findings by examining whether results reported by the original authors can be reproduced. Second, we test the \emph{replicability} of executable codes on two datasets: a dataset similar to the one used in the original paper and a new dataset built by manually annotating tables in six scientific domains. Our main contributions are summarized as follows:

\begin{enumerate}
    \item We perform a study on AI reproducibility and replicability based on state-of-the-art TSR methods and identify reproducible and replicable papers under certain conditions.
    \item We build a new, manually-annotated dataset, GenTSR, representing digital tables in papers from six scientific domains and demonstrate that the dataset is more challenging than widely adopted benchmarks such as ICDAR 2013 and ICDAR 2019, on the TSR task.  
    \item We observe possible causes of non-reproducibility and non-replicability of AI papers based on our experiments. 
\end{enumerate}

\section{Related Work}
Concerns about reproducibility in computer science have been studied in the context of computer systems, e.g., \cite{collberg2016repeatability}, software engineering, e.g., \cite{liu2021se}, and recently on artificial intelligence (AI). e.g., \cite{gundersen2018state,raff2019step}. 
Recent efforts have characterized the reproducibility of AI papers using automatic verification or meta-level information. Pimentel et al. \cite{jupyter-reproducibility} conducted an extensive study on over 1 million Jupyter notebooks from GitHub and found that only 24.11\% executed without errors and only 4.03\% produced the same results. Kamphius et al. conducted a large-scale study of reproducibility on BM25 scoring function variants \cite{kamphuis2020bm25}. Seibold et al. \cite{seibold2021computational} investigated the reproducibility of analyses of longitudinal data associated with 11 articles published in PLOS ONE after contacting original authors. 
Prenkaj \cite{Prenkaj} compared several deep methods for trajectory forecasting on different datasets to provide insight into the actual novelty, reliability, and applicability of available methods. Salsabil et al. \cite{salsabil2022scik} proposed a hybrid classifier to automatically extract open-access datasets and software from scientific papers. Gunderson et al. \cite{gundersen2018state} investigated 400 AI papers, and found that none contain documentation of published experiments, methods, and data altogether.

Although the work referenced above have highlighted the importance of including codes and data alongside published findings, studies were limited to meta-level indicators. There are a handful of studies that directly compare reproduced results of AI algorithms with published results. For example, Olorisade et al. \cite{olorisade2017reproducibility} attempted to directly reproduce 6 AI papers using the codes and datasets from the original papers, reporting inconsistent results with published findings.  
Raff \cite{raff2019step} directly compared the reported results of 255 AI papers without using the code from the original papers, and found out that 162 papers were at least 75\% consistent with reported results while 93 were not.

Lack of transparency and reproducibility is particularly critical given the standard for AI papers to evaluate performance of proposed methods against baselines. 
Such problems have been found in AI research on machine learning analyses on clinical research \cite{stevens2020recommendations} and deep metric learning \cite{musgrave2020metric}. \emph{Replication} studies that provide side-by-side comparisons between AI papers addressing the same topic are rarely conducted. 
Our work fills this gap by directly comparing the implementations of TSR methods with reported results, and testing the replicability of these methods on new datasets. We chose TSR because recently, many learning-based methods on this task have been proposed and reportedly achieved high performance but not all of them were evaluated on the same datasets. Standard benchmarks are available in open competitions, i.e., ICDAR 2013 \cite{gobel2013icdartable} and ICDAR 2019 (Task~B2) \cite{gao2019icdartable}. Although the datasets were created 6 years apart, both resemble \emph{generic} tables in a variety of documents, including government documents, scientific journals, forms, and financial statements. Scientific tables, on the other hand,  
usually contain precise measurements of experimental or analytical results. Compared with other types of tables, scientific tables are more heterogenous, with complex and freestyle structures. Therefore, for our replication study, we built a separate evaluation benchmark using tables extracted from six scientific domains to challenge existing runnable TSR algorithms. 
Our work is different from a typical survey paper in that our focus is not on outlining the proposed algorithms but on testing the reproducibility and replicability of state-of-the-art TSR algorithms.

\begin{figure}
    \centering
    \includegraphics[scale = 0.65]{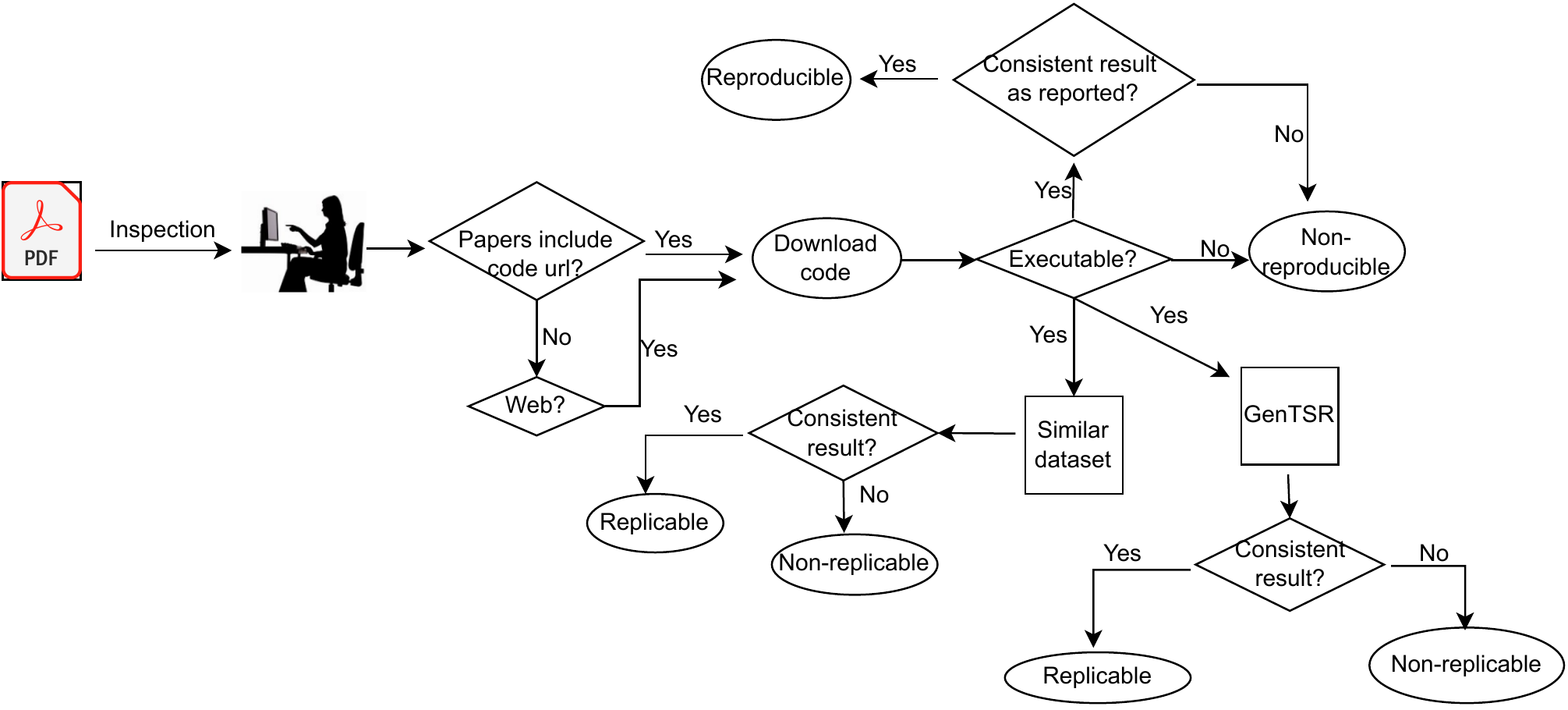}
    \caption{Workflow of our study.}
    \label{pub-exp}
\end{figure}

\section{Methods}
Prior work by Tatman et al. proposed a taxonomy of reproducibility for AI,  
namely, 
``low'', ``medium'', and ``high'' based on the availability of code, data, and adequate documentation of the experimental environment. This taxonomy is not directly applicable to our work because we not only verify the availability of code and data but also execute the code and compare with reported results. Figure~\ref{pub-exp} illustrates the workflow of our study. The process is summarized below. 
\begin{enumerate}
    \item {\bf Sample Selection.} TSR papers were selected by searching ``table structure recognition'' as keywords on Google Scholar. Results were filtered to include papers published after 2017, in which the proposed methods accepted documents or table images as input. We downloaded 25 TSR papers from the conference websites, and we selected 16 candidate papers which use deep-learning based methods as our final sample. 

    \item {\bf Meta-level Study.} We conducted meta-level study by inspecting each paper and determining whether the authors included URLs linking to source codes and datasets. If no URLs were found, we attempted to find open access codes and datasets by searching author names and framework names on Google. We bookmarked code and data repositories. 
   
    \item {\bf Local Deployment.} We downloaded the data and source codes and deployed them in local computers by following the instructions in the original paper or on the code repositories.
    \item {\bf Reproducibility Tests.} We attempted to execute the codes using default settings and labeled each paper into one of three categories. 
    \begin{enumerate}
      \item \emph{Reproducible:} The source code was executed without errors and the results were consistent with the reported results within a deviation of an absolute F1-score of 10\% below or above the reported results.
      \item \emph{Partially-reproducible:} The source code was executed without errors and the results were better than the reported results by more than an absolute F1-score of 10\% deviation from the paper.
      \item \emph{Non-reproducible:} Otherwise.
    \end{enumerate}

    \item {\bf Replicability Tests on a Similar Dataset.} We tested executable TSR methods on a different but similar benchmark dataset. If results are consistent with the reported results within a deviation of an absolute F1-score of 10\% under certain conditions, the paper was labeled as ``conditionally replicable'' with respect to this dataset. 
    \item {\bf Replicability Tests on a New Dataset.} We tested executable TSR methods on a new dataset. If results are consistent within a deviation of 10\% absolute F1-score, the paper was labeled as ``conditionally replicable'' with respect to the new dataset. 
\end{enumerate}

We create a separate virtual environment for each TSR method to avoid incompatibility issues. Then, we execute the code of each TSR algorithm to reproduce or replicate the reported F1 scores.
Because we focus on reproducing results presented in original papers, we used the pre-trained models released by the authors. One important parameter that may affect the TSR result is Intersection over Union (IoU), which quantifies the percentage overlap between object regions provided by the ground truth and predicted by the model. In practice, IoU is measured by dividing the number of common pixels between the ground-truth bounding boxes and predicted bounding boxes by the total number of pixels across both bounding boxes. The IoU threshold defines the criterion of whether the predicted bounding boxes match the ground truth bounding boxes. The matching between two bounding boxes is counted if the predicted IoU is larger than the threshold. 

For our reproducibility tests, we evaluate an executable model on the same datasets used in the original paper, either ICDAR~2013 or ICDAR~2019. If a paper used both ICDAR-2013 and ICDAR 2019 datasets, then we chose ICDAR-2019 because it contains more challenging tables. If a paper used neither dataset, then we used the dataset used in the original paper. For the replicability tests on a similar dataset, we evaluated each executable model on the alternate dataset of the two benchmarks, e.g., if ICDAR 2013 was used in the original paper, we use ICDAR 2019. For the second replicability test, we evaluated the model on a new dataset called GenTSR (introduced below). We compute F-scores at five IoU thresholds $0.5$, $0.6$, $0.7$, $0.8$, and $0.9$. 

For reproducibility tests, we define the discrepancy $\Delta$ as the absolute difference between the F1-score obtained by our reproduction $F1(R_0)$ and the F1-score reported in the original paper $F1(O)$, i.e.,  $\Delta_0={F1(R_0)}-{F1(O)}$. For  replicability tests, discrepancy $\Delta$ is defined as F1-score of our replication $F1(R_x)$ and F1-score of our reproduction  $F1(O)$, i.e., $\Delta_x={F1(R_x)}-{F1(R_0)}$. We do not compare replicability against the original F1-score to ensure that compared results are obtained in exactly the same setting. 

\section{Data}
We use two standard benchmarks and GenTSR, our manually-annotated dataset. 

\paragraph{\bf ICDAR~2013.} This dataset, released for the table competition by ICDAR 2013, was used in 8 papers out of 16 papers (Table~\ref{TSR}). ICDAR 2013 consists of 238 document pages in PDF format crawled from European and US government websites, out of which 128 documents include tables. We did not use the original ICDAR 2013 data  as our ground truth because it consists of born-digital PDFs and all the TSR models we surveyed accept either documents or table images as input. Therefore, we cropped the tables from the PDFs based on the labeled coordinates and automatically adjust its annotations accordingly. 

\paragraph{\bf ICDAR~2019.} This dataset, released for the Competition on Table Detection and Recognition (cTDaR) organized by ICDAR 2019, was used in 6 out of 16 papers we surveyed. The cTDaR competition includes two datasets including the modern and historical tables, respectively. The modern dataset contains 100 samples from scientific papers, forms, and financial documents, and the historical dataset includes images from hand-written accounting ledgers, and train schedules. Our experiments adopt the modern table dataset used in Track B2 (TB2).

\begin{table*}[ht]
\begin{adjustbox}{width=\textwidth,center}
\centering
\begin{threeparttable}
\caption{A summary of TSR papers used in our study, their properties, and direct reproducibility labels (``Rep'' column). The columns labeled ``Data'' and ``Code'' indicate whether datasets and codes are publicly available. A dash (``-'') means the resource is not available.}

\label{TSR}
  \begin{tabular}{c|c|c|c|c|c|c|c|c}
    \toprule
    \textbf{Reference} & \textbf{Year} & \textbf{Model Name}   & \textbf{Venue} & \textbf{Training data} & \textbf{Original Eval. data} & \textbf{Data} & \textbf{Code} & {\bf Rep}\\
    \midrule
    \textbf{Schreiber et al. }\cite{DeepDsrt} & 2017 & {DeepDeSRT} & {ICDAR} & {Marmot} & {Marmot} & {-} & {-} & {NR}\\  [5pt]
     \midrule
      \textbf{Siddiqui et al. }\cite{deeptabstr} & {2019} & {DeeptabSTR}  & {ICDAR} & {TabStructDB} & {ICDAR 2013}& {-} & {-} &{NR}\\  [5pt]
 \midrule
 \textbf{Xue et al.}\cite{ReS2TIM}\tnote{1}  & {2019} & {Res2TIM}  & {ICDAR} & \makecell{CMDD + \\ ICDAR 2013} & {ICDAR 2013} & {\checkmark}& {\checkmark} & {R} \\  [5pt]
 \midrule
\textbf{Qasim.}\cite{qasim2019rethinking} \tnote{2}& 2019 & TIES-2.0  & {ICDAR} & Synthetic data & Synthetic data & {\checkmark}& {\checkmark} & \makecell{NR}\\  [5pt]
 \midrule
\textbf{Tensmeyer et al.}\cite{splerge}\tnote{4}   & 2019 & SPLERGE & {ICDAR} & \makecell{Web-screaped PDFs \\ + ICDAR 2013} & ICDAR 2013 & {-}  & {\checkmark} & {NR} \\  [5pt]
\midrule

\textbf{Prasad et al.}\cite{cascadetabnet}\tnote{5} & 2019 & \makecell{Cascade\\TabNet}  & {CVPR} & Marmot + ICDAR 2019 & \makecell{ICDAR 2019 \\ Track-B2} & {\checkmark} & {\checkmark} & {PR}\\  [5pt]
\midrule

 \textbf{Hashmi et al.}\cite{guided} & 2019 & {No name}  & {CVPR} & {TabStructDB} & {ICDAR 2013} & {-} & {-} &{NR}\\  [5pt]
 \midrule
\textbf{Khan et al.}\cite{GRU-TSR}\tnote{3} & {2020} & {No name}  & {ICDAR} & {UNLV} & {ICDAR 2013}& {-}  & {\checkmark} & \makecell{NR}\\  [5pt]
\midrule

\textbf{Raja et al.}\cite{TabStructNet}\tnote{6}& 2020 & \makecell{TabStruct\\Net}  & {ECCV} & SciTSR & UNLV & {\checkmark} & {\checkmark} & \makecell{NR}\\  [5pt]
\midrule
\textbf{Fischer et al.}\cite{Multi-Type-TD-TSR}\tnote{7} & 2021 & \makecell{Multi-Type\\-TD-TSR}  & {KI} & ICDAR 2019 & \makecell{ICDAR 2019 \\ Track-B2} & {\checkmark} & {\checkmark} & {R} \\ [5pt]  
\midrule
\textbf{Xue et al.}\cite{TGRNet}\tnote{8}  & 2021 & TGRNet  & {ICCV} & TableGraph & ICDAR 2019 & {\checkmark} & {\checkmark}  & {R}\\  [5pt]
 \midrule
 
  \textbf{Qiao et al.}\cite{qiao2021lgpma}\tnote{10}  & 2021 & {LGPMA}   & {ICDAR} & \makecell{PubTabNet + SciTSR +  \\ ICDAR 2013} & {PubTabNet} & {\checkmark} & {\checkmark} & \makecell{NR}\\  [5pt]
 \midrule
 
  \textbf{Lee et al.}\cite{Graph-based-TSR-lee}\tnote{11}  & 2021 & \makecell{Graph-based\\-TSR}  & {MTA} & ICDAR 2019 & ICDAR 2019 & {\checkmark} & {\checkmark} &  {R}\\  [5pt]
 \midrule

 \textbf{Zheng et al.}\cite{GTE}  & 2022 & {GTE}  & {WACV} & {PubTabNet} & \makecell{ICDAR 2013 + \\ ICDAR 2019} & {-} & {-} & NR\\  [5pt]
 \midrule
 
  \textbf{Jain et al.}\cite{DBLP:conf/esann/JainPSV21}\tnote{9}  & 2022 & {TSR-DSAW}   & {ESANN} & {PubTabNet} & {ICDAR 2013} & {-} &{-}  & \makecell{NR}\\  [5pt]
 \midrule

   \textbf{Li et al. }\cite{li2021rethinking}\tnote{12} & 2021 & {No name} & {ICDAR} & {PubTabNet} & \makecell{ICDAR 2019 + \\ unlv} & {-} & {\checkmark} &  \makecell{NR}\\  [5pt]
     \bottomrule

  \end{tabular}
  
\begin{tablenotes}
\item Notes: NR: non-reproducible. R: Reproducible. PR: Partially-reproducible. CMDD:Chinese Medical Document Dataset. Rep: Reproducibility\\
\item [1] \url{ https://github.com/xuewenyuan/ReS2TIM/}\\
\item [2] \url{https://github.com/shahrukhqasim/TIES-2.0}\\
\item [3] \url{https://github.com/saqib22/Table-Structure\_Extraction-Bi-directional-GRU/}\\
\item [4] \url{ https://github.com/pyxploiter/deep-splerge/}.\\
\item [5] \url{ https://github.com/DevashishPrasad/CascadeTabNet/}\\
\item [6] \url{https://github.com/sachinraja13/TabStructNet}\\
\item [7] \url{ https://github.com/Psarpei/Multi-Type-TD-TSR/}\\
\item [8] \url{ https://github.com/xuewenyuan/TGRNet/}\\
\item [9] \url{https://github.com/arushijain45/TSR-DSAW/}\\
\item [10] \url{https://github.com/hikopensource/DAVAR-Lab-OCR/tree/main/demo/table\_recognition/lgpma/}\\
\item [11] \url{ https://github.com/ejlee95/Graph-based-TSR/}\\
\item [12] \url{ https://github.com/L597383845/row-col-table-recognition/}
\end{tablenotes}

\end{threeparttable}
\end{adjustbox}
\end{table*}

\paragraph{\bf GenTSR.} This dataset consists of 386 table images obtained from research papers in six scientific domains, including three STEM (Chemistry, Biology, and Materials Science) and three non-STEM domains (Economics, Developmental Studies, and Business). The format of GenTSR is consistent with ICDAR 2019. The numbers of tables in each domain are 30 (Chemistry), 43 (Economics), 7 (Developmental Studies), 68 (Biology), and 208 (Materials Science). These tables were manually annotated by two graduate students independently using the VGG Image Annotator (VIA) \cite{dutta2019via}. VIA is open-source software for annotating images, videos, and audio. We drew rectangular bounding boxes around text content in a table cell and provided properties including ``start-row'', ``start-col'', ``end-row'', and ``end-col'' as labels. We followed the same schema as ICDAR 2019 dataset. We obtained a Cohan's $\kappa=0.73$, indicating a substantial agreement between the two annotators \cite{mchugh2012kappa}. The two annotators then discussed until they agreed on the remaining table cells that they initially did not agree with each other. 

\begin{table*}
 \caption{Three datasets used in our study. }
    \centering
    \begin{tabular}{c|c|c|c|c}
    \toprule
      {\bf Data}  & {\bf \#tables} & {\bf \#cells} & {\bf \#row ranges} & {\bf \#column ranges}  \\
    \midrule
    ICDAR-2013  & 158 & 14,278 & 2 - 58 & 2 - 13 \\
    ICDAR-2019 & 100 & 5,132 & 2 - 39 & 1 - 15 \\
    GenTSR & 386 & 19,914 & 2 - 62 & 1 - 16 \\
   
    \bottomrule
  
    \end{tabular}
   
    \label{tab:prop}
\end{table*}

\section{Experiment Results}
We performed all experiments using two computing environments namely, a Linux server with Intel Silver CPU, Nvidia GTX 2080 Ti and Google Colaboratory platform with P100 PCIE GPU of 16 GB GPU memory. It took approximately 12 hours to reproduce the 5 papers that made their codes and data available. Specifically, it took about 6 hours to reproduce TGRNet and Res2TIM, and approximately 2 hours each for CascadeTabNet, Graph-based-TSR, and Multi-Type-TSR. The replication experiment took about 18 hours (excluding the time to create the GenTSR dataset) even though all necessary packages used for each method had already been installed when conducting reproducibility experiment. This was due to the relatively large size of replication data and multiple replication attempts to cross-check results.

\begin{figure}[h!]
    \centering
    \includegraphics[scale=0.40, center]{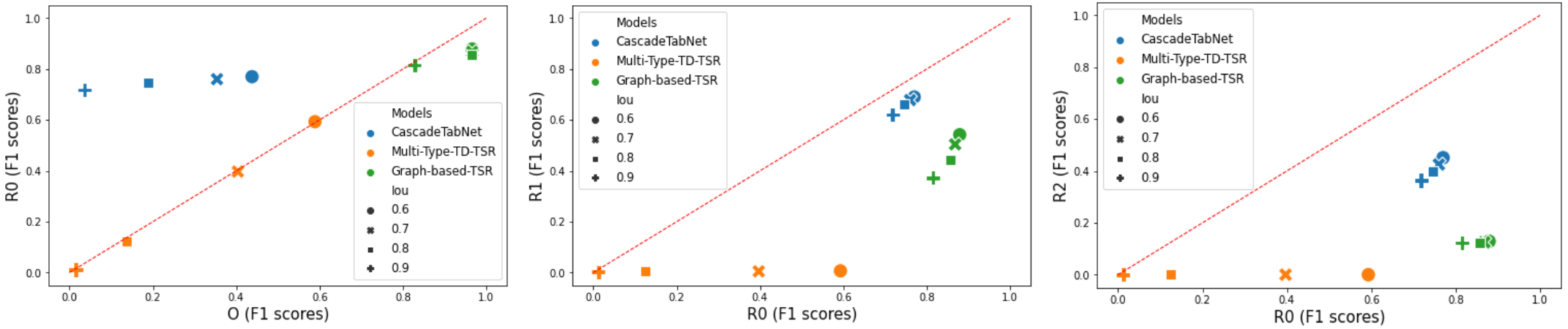}
    \caption{The comparison of the original (O), reproducibility (R0), replicability on similar data (R1), and replicability on GenTSR (R2). The F1-scores of R2 are obtained by averaging the F1-scores across all domains for each IoU. SPLERGE was excluded because its results were not reproducible.}
     \label{fig:gen}
\end{figure}

\begin{figure}[h!]

    \centering
    \includegraphics[scale=0.58]{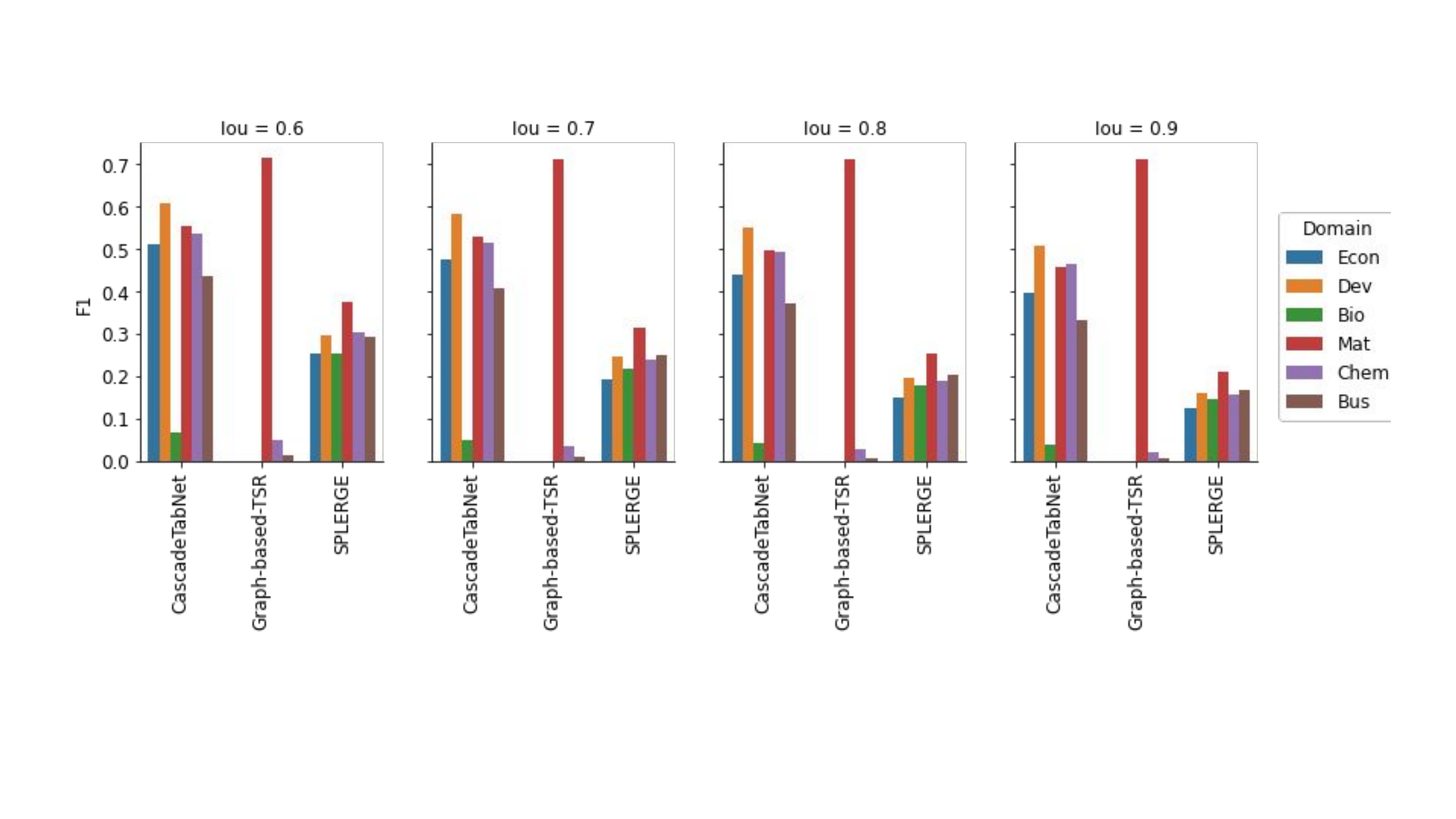}
     \caption{The \emph{replicability test} results of executable TSR model with respect to individual domains in GenTSR. Econ: Economics, Dev: Developmental Studies, Bio: Biology, Mat: Material Science, Chem: Chemistry, Bus: Business. Multi-Type-TD-TSR (not shown) obtains recall scores of zero across all the domains.}
     \label{fig:domain}
\end{figure}

We answer the following research questions (RQs) using meta-level survey results and reproducibility and replicability experiment results. The reproducibility results are tabulated in Table~\ref{tab:comparison}. The replicability results are illustrated in Figure~\ref{fig:gen} and Figure~\ref{fig:domain}.

\paragraph{\textbf{RQ1: What is the data and code accessibility of TSR papers we sampled?}} 
Out of 16 papers we surveyed, 8 papers made their source code and data publicly available, 3 papers made only the codes available, and 5 papers did not provide either codes or data, making it difficult to validate the results of these methods without private communication with the original authors (Table~\ref{TSR}). 

\paragraph{\textbf{RQ2: Are the accessible methods executable without contacting the original authors?}}
Out of 11 papers with accessible data or codes, the codes of 5 papers were executable without contacting the original authors. The source codes of 6 papers were not executable \cite{qasim2019rethinking,GRU-TSR,TabStructNet,qiao2021lgpma,DBLP:conf/esann/JainPSV21,li2021rethinking}. The code of one paper (TGRNet; \cite{TGRNet}) was executable after we contacted the original authors. The reason was that the absolute paths to the evaluation data files in the original code were hard-coded. Therefore, the program could not find data files after they are transferred to a different environment. To evaluate the models, we wrote a script to replace the paths and the source codes could be executed. The source codes of the 6 papers were not executable due to multiple reasons such as dependency issues, errors in code, pretrained models not being released, or the absence of implementation in the authors' GitHub directory.  

\paragraph{\textbf{RQ3: What is the status of reproducibility based on our criteria?}}
The status of reproducibility varies significantly depending on many factors. As shown above, most papers were labeled non-reproducible because they do not provide datasets, codes, or executable codes. However, most papers with executable codes were labeled reproducible under our criteria. Specifically, 4 out of the 6 executable TSR methods were labeled reproducible, 1 paper was labeled partially-reproducible, and 1 paper was labeled not-reproducible. The case studies are below.

\begin{itemize}
  \item Lee et al. \cite{Graph-based-TSR-lee} used only 19 document images with border lines from the ICDAR~2019 TB2 dataset to evaluate the Graph-based-TSR method. Therefore, we evaluated this method on the same 19 images. Table~\ref{tab:comparison} indicates that the reproduction results are in general consistent with the reported results, with discrepancies $0.087\leq\Delta_0\leq0.130$ depending on the IoU. 
  \item The CascadeTabNet method was originally evaluated on 100 modern tables in ICDAR 2019. Surprisingly, our experiment on CascadeTabNet obtained higher F1-scores than the reported results by up to $0.682$ at ${\rm IoU}=0.9$. 
  \item The Multi-Type-TD-TSR method was evaluated on 162 tables from ICDAR-2019 TB2. The experiment results are consistent with the reported results with a discrepancy $\Delta_0\leq0.013$. 
  \item TGRNet and ReS2TIM are both consistent with the reported results with a discrepancy $\Delta_0\leq0.003$. The authors of these two methods used only IoU = 0.5 to allow more cell boxes to be predicted.

  \item The SPLERGE method was originally evaluated on 34 randomly selected tables using ICDAR 2013 dataset but this dataset was not made publicly available. Thus, the SPLERGE method was marked ``Non-reproducible''.
\end{itemize}

\paragraph{{\bf RQ4: What reasons caused results to be not reproducible?}}
We identified several major reasons that caused the results to be not-reproducible. 
\begin{itemize}
\item \textbf{Data and code availability:} This is the top reason that caused most papers to be non-reproducible. However, most papers \emph{with executable codes} are identified as reproducible. 

Several non-reproducible papers have authors affiliated with the industry, which may impose intellectual property restrictions, e.g.,\cite{GTE,DeepDsrt,deeptabstr,qasim2019rethinking,GRU-TSR,guided}.

\item \textbf{Portability:} Agile software engineering may develop software packages that are not portable when transferred to other platforms, e.g., \cite{TabStructNet}. 
  
\item \textbf{Documentation:} This occurs when researchers do not provide detailed instructions or explanations to execute their codes, e.g., \cite{DBLP:conf/esann/JainPSV21}. 

\item \textbf{Dependency and compatibility issues:} Software that relies on outdated dependencies can become prohibitive obstacles to reproducing reported results. Certain software did not provide the dependency version, making it extremely difficult or even impossible to find and install the right dependency, e.g., \cite{qiao2021lgpma}.

\item \textbf{Data and code durability:} This occurs when the data and codes used in the original paper are updated after published results. Thus, a better result than what was reported may be obtained after executing the updated codes and data, thereby making it difficult to validate original reported results.
\end{itemize}

\begin{table*}[t]
 \caption{The \emph{reproducibility test} results of executable TSR models at different IoU thresholds. Data: the \emph{original dataset}. SPLERGE does not provide evaluation data. }
    \centering
    \begin{tabular}{c|c|c|c|c|c}
    \toprule
      {\bf TSR Model}  & {\bf Data} & {\bf IoU} & {$F1(O)$} & {$F1(R_0)$} & $\Delta_0$  \\
    \midrule
    CascadeTabNet & ICDAR 2019 & 0.6 & 0.438 & 0.770 & 0.332 \\
    CascadeTabNet & ICDAR 2019 & 0.7 & 0.354 & 0.760 & 0.406 \\
    CascadeTabNet & ICDAR 2019 & 0.8 & 0.190 & 0.745 & 0.555 \\
    CascadeTabNet & ICDAR 2019 & 0.9 & 0.036 & 0.718 & 0.682 \\
   
    \midrule
    Multi-Type-TD-TSR & ICDAR 2019 & 0.6 & 0.589 & 0.593 & 0.004 \\
    Multi-Type-TD-TSR & ICDAR 2019 & 0.7 & 0.404 & 0.397 & -0.007 \\
    Multi-Type-TD-TSR & ICDAR 2019 & 0.8 & 0.137 & 0.124 & -0.013 \\
    Multi-Type-TD-TSR & ICDAR 2019 & 0.9 & 0.015 & 0.012 & -0.003 \\
    \midrule
    Graph-based-TSR & ICDAR 2019 & 0.6 & 0.966 & 0.879 & -0.087 \\
    Graph-based-TSR & ICDAR 2019 & 0.7 & 0.966 & 0.868 & -0.098 \\
    Graph-based-TSR & ICDAR 2019 & 0.8 & 0.966 & 0.856 & -0.110 \\
    Graph-based-TSR & ICDAR 2019 & 0.9 & 0.828 & 0.815 & -0.130 \\
    \midrule
    TGRNet & ICDAR 2013 & 0.5 & 0.667 & 0.670 & 0.003 \\
    \midrule
    ReS2TIM & ICDAR 2013 & 0.5 & 0.174 & 0.174 & 0.000 \\
    \midrule
    SPLERGE & ICDAR 2013 & 0.5 & 0.953 & - & - \\
    \bottomrule
    \end{tabular}
   
    \label{tab:comparison}
\end{table*}

\begin{table*}[h!]
 \caption{The \emph{replicability test} results of executable TSR models at different IoU thresholds. Data: the \emph{similar dataset}. TGRNet and ReS2TIM do not allow inference on a custom dataset.}
    \centering
    \begin{tabular}{c|c|c|c|c|c}
    \toprule
      {\bf TSR Model}  & {\bf Data} & {\bf IoU} & {$F1(R_0)$} & {$F1(R_1)$} & $\Delta_1$  \\
    \midrule
    CascadeTabNet & ICDAR 2013 & 0.6 & 0.770 & 0.690 & -0.080\\
    CascadeTabNet & ICDAR 2013 & 0.7 & 0.760 & 0.678 & -0.082 \\
    CascadeTabNet & ICDAR 2013 & 0.8 & 0.745 & 0.661 & -0.084 \\
    CascadeTabNet & ICDAR 2013 & 0.9 & 0.718 & 0.621 & -0.097 \\
   
    \midrule
    Multi-Type-TD-TSR & ICDAR 2013 & 0.6 & 0.593 & 0.007 & -0.586 \\
    Multi-Type-TD-TSR & ICDAR 2013 & 0.7 & 0.397 & 0.005 & -0.392 \\
    Multi-Type-TD-TSR & ICDAR 2013 & 0.8 & 0.124 & 0.004 & -0.120 \\
    Multi-Type-TD-TSR & ICDAR 2013 & 0.9 & 0.012 & 0.003 & -0.009 \\
   
    \midrule
    Graph-based-TSR & ICDAR 2013 & 0.6 & 0.879 & 0.542 & -0.337 \\
    Graph-based-TSR & ICDAR 2013 & 0.7 & 0.868 & 0.504 & -0.364 \\
    Graph-based-TSR & ICDAR 2013 & 0.8 & 0.856 & 0.444 & -0.412 \\
    Graph-based-TSR & ICDAR 2013 & 0.9 & 0.815 & 0.373 & -0.442 \\
    \midrule
    TGRNet & ICDAR 2019 & 0.5 & 0.670 & - & - \\
    \midrule
    ReS2TIM & ICDAR 2019 & 0.5 & 0.174 & - & - \\
    \midrule
    SPLERGE & ICDAR 2019 & 0.5 & - & 0.121 & - \\
    \bottomrule
 
    \end{tabular}
   
    \label{tab:generalization}
\end{table*}

\paragraph{\textbf{RQ5: What is the status of replicability with respect to a similar dataset?}}
To answer this question, we evaluated each executable TSR method on a similar dataset. Here, we compare against the reproducibility experiment results instead of the original results to ensure the results to be compared are obtained in exactly the same setting. 

Table~\ref{tab:generalization} indicates the F1-scores of most methods were reduced by various levels depending on the IoU thresholds. In particular, the F1 of Graph-based-TSR decreases by $0.337$. The F1 of Multi-Type-TD-TSR decreases by $0.009$ to $0.586$. The performance of CascadeTabNet decreases marginally, exhibiting better replicability. We could not replicate the results of ReS2TIM and TGRNet because they do not allow inference on an alternative dataset. Also, we did not obtain the discrepancy $\Delta_1$ for the SPLERGE method since it was not reproducible. Thus, out of the 6 methods that were either executable or reproducible, only 2 papers (CascadeTabNet and MUlti-Type-TD-TSR) were replicable under certain IoUs (CascadeTabNet on IoU from 0.6 to 0.9; Multi-Type-TD-TSR on IoU=0.9).

\paragraph{\textbf{RQ6: What is the status of replicability with respect to the new dataset?}}
We test the replicability of each executable TSR model using GenTSR containing tables in six scientific domains. Similar to \emph{RQ5.}, we compare against the reproducibility experiment results using the 10\% threshold defined above. The results shown in Figure~\ref{fig:gen} and Figure~\ref{fig:domain} indicate that none of the 4 methods that allow inference on custom data \cite{cascadetabnet,Graph-based-TSR-lee,Multi-Type-TD-TSR,splerge} was replicable with respect to the GenTSR dataset, under a threshold of 10\% absolute F1-score. Figure~\ref{fig:domain} also demonstrates that the performance of these methods varies significantly in scientific domains. Specifically, the CascadeTabNet achieved much higher F1-scores on five domains than biology. SPLERGE achieves comparable F1-scores in all domains. Graph-based-TSR performs remarkably well in Material Science but poorly in all other domains.

\section{Discussion}
\paragraph{Reproducibility} In reproducibility experiments, we were unable to produce exactly the same results reported in most papers. The discrepancies may be due to random factors, e.g., initialization, but certain discrepancies are too large to be explained by random factors. Investigating the reasons is beyond the scope of this paper, but the results suggest we define reproducibility using quantifiable criteria associated with thresholds. The exact criteria will differ depending on the results and the reproducer's needs. In addition, we obtained an interesting result in which the reproduced results were significantly better than the reported result (Table~\ref{tab:comparison}). Assuming both original and reproduced results are correct, the improvement could be attributed to the new versions of the codes and/or data. If that is the case, this poses another question of how long can reproducibility be preserved. 

\paragraph{Replicability is more challenging and data dependent.} One requirement of replicability is that the original code is not only executable but also configurable, allowing users to test on different datasets. In our experiments, two methods did not allow inference on different datasets. In addition, the replicability performance could change the ranks of methods. For example, the Graph-based-TSR was the best in terms of the original and reproduced F1-scores, but it underperformed CascadeTabNet in two replicability tests (Table~\ref{tab:generalization} and Figure~\ref{fig:gen}). This is likely to be caused by the nature of the model and the training process. The exact reason requires detailed ablation analysis and model surgery. 
In addition, only CascadeTabNet obtained reasonably consistent results using a similar dataset (Table~\ref{tab:generalization}). Using the new dataset, CascadeTabNet achieved a lower $F1(R_2)$ compared with $F1(O)$ and $F1(R_1)$. Figure~\ref{fig:domain} indicates that the performance exhibits a strong domain dependency. The decreased performance as seen in the new dataset indicates that TSR on scientific tables is still an unsolved problem and state-of-the-art methods still have a large space to improve. 

\paragraph{Potentially non-replicability causes: evaluation bias} The replicability test results also indicate that the evaluation data of several TSR models may not be diverse enough. For example, the Graph-based-TSR was evaluated on only tables with borders. In contrast, models that exhibit better robustness tend to be evaluated originally on more challenging datasets. For example, CascadeTabNet, which obtained the best replicability results was evaluated on ICDAR 2019 Task~B2 which was more challenging and diverse compared to ICDAR~2013 and other small benchmarks. 

\begin{figure}
\centering
\includegraphics[width=0.8\textwidth]{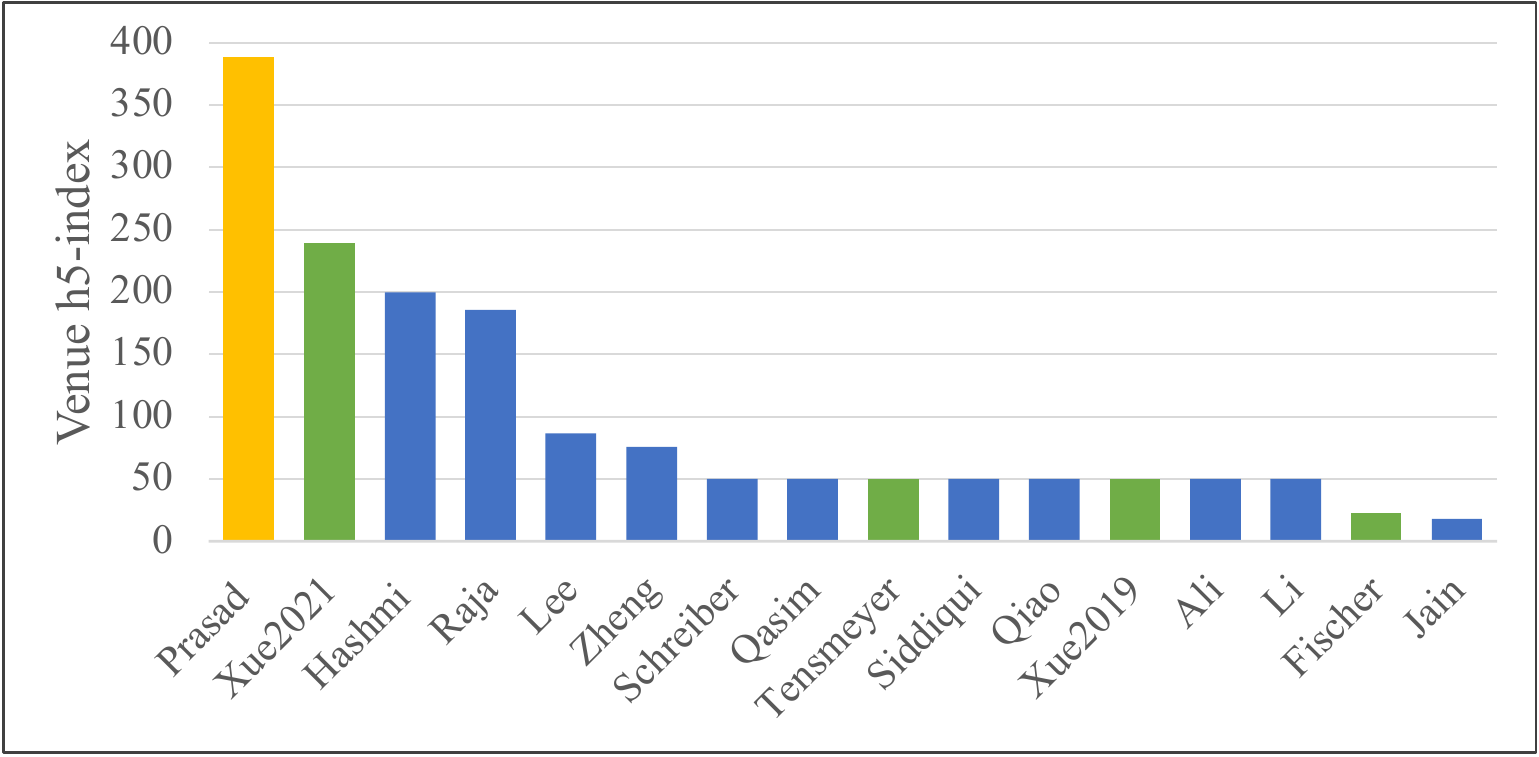}
\caption{The relationship between reproducibility and venue ranking, characterized by h5-index from Google Scholar. The h5-indices of ESANN (Jain) and KI (Fischer) are not available, so we used h-indices as surrogates, obtained from research.com and resurchify.com, respectively.\label{fig:venueindex}}
\end{figure}

\paragraph{Reproducibility and venue ranking} We inspected the relationship between reproducibility and venue ranking, which is characterized by the h5-index obtained in December, 2022. Created by Google Scholar, the calculation of h5-index is similar to h-index. H5-index is defined as the largest number $h$ such that $h$ articles published in the past 5 years have at least $h$ citations each. Figure~\ref{fig:venueindex} shows the papers we studied and color-coded by reproducibility. It indicates that reproducibility is not necessarily associated with venue ranking. 

\paragraph{Limitations} One limitation of our study is the relatively small sample size. Therefore the conclusions we draw may not directly be applicable to other AI tasks and domains. However, the way we selected the papers allowed us to focus on more papers on one topic and perform a side-to-side comparison between different methods. Our threshold of 10\% absolute F1-score is also a very lenient threshold. Under a more strict threshold, fewer papers would be identified as reproducible.

\section{Conclusion}
This work presents a study of reproducibility and replicability considering 16 recently-published papers on table structure recognition. We attempted to directly reproduce the results reported in the original papers. We then tested executable methods on an alternate benchmark dataset similar to the one used in the original paper as well as a new dataset of modern scientific tables extracted from six  domains. 

Under our criteria, most (12 out of 16) papers we examined were not fully reproducible. Only 2 papers \cite{cascadetabnet,Multi-Type-TD-TSR} were identified as conditionally replicable on a similar  dataset, and none of the papers was identified as replicable with respect to the new dataset. Using a relatively small but focused dataset, our study reveals several challenges of reproducing and replicating methods proposed for the TSR task. Our study suggests that reproducibility should be defined under certain criteria with quantifiable thresholds and replicability is data-dependent. We also found that papers published in high-tier venues (characterized by h5-index) are not necessarily reproducible. The new dataset GenTSR can be used as ground truth for building more robust TSR models. Future work will investigate replicability at the model level. We suggest that this work provides evidence that infrastructure is needed for researchers to report RR\&G of experiments.  

\bibliographystyle{unsrt}
\bibliography{references}

\begin{thebibliography}{10}

\bibitem{fanelli2017pnas}
Daniele Fanelli.
\newblock Opinion: Is science really facing a reproducibility crisis, and do we
  need it to?
\newblock {\em Proceedings of the National Academy of Sciences},
  115(11):2628--2631, 2018.

\bibitem{baker20161500}
Monya Baker.
\newblock 1,500 scientists lift the lid on reproducibility.
\newblock {\em Nature}, 533(7604):452--454, 2016.

\bibitem{camerer2018evaluating}
Colin~F. Camerer, Anna Dreber, Felix Holzmeister, Teck-Hua Ho, J{\"u}rgen
  Huber, Magnus Johannesson, Michael Kirchler, Gideon Nave, Brian~A. Nosek,
  Thomas Pfeiffer, Adam Altmejd, Nick Buttrick, Taizan Chan, Yiling Chen, Eskil
  Forsell, Anup Gampa, Emma Heikensten, Lily Hummer, Taisuke Imai, Siri
  Isaksson, Dylan Manfredi, Julia Rose, Eric-Jan Wagenmakers, and Hang Wu.
\newblock Evaluating the replicability of social science experiments in nature
  and science between 2010 and 2015.
\newblock {\em Nature Human Behaviour}, 2(9):637--644, 2018.

\bibitem{olorisade2017reproducibility}
Babatunde~Kazeem Olorisade, Pearl Brereton, and Peter Andras.
\newblock Reproducibility of studies on text mining for citation screening in
  systematic reviews: Evaluation and checklist.
\newblock {\em Journal of Biomedical Informatics}, 73:1--13, 2017.

\bibitem{raff2019step}
Edward Raff.
\newblock A step toward quantifying independently reproducible machine learning
  research.
\newblock {\em Advances in Neural Information Processing Systems}, 32, 2019.

\bibitem{goodman2016}
Steven~N Goodman, Daniele Fanelli, and John~PA Ioannidis.
\newblock What does research reproducibility mean?
\newblock {\em Science translational medicine}, 8(341):341ps12--341ps12, 2016.

\bibitem{national2019reproducibility}
National Academics.
\newblock {\em Reproducibility and Replicability in Science}.
\newblock National Academies Press, 2019.

\bibitem{nosek2021replicability}
Brian~A. Nosek, Tom~E. Hardwicke, Hannah Moshontz, Aur\'{e}lien Allard,
  Katherine~S. Corker, Anna Dreber, Fiona Fidler, Joe Hilgard, Melissa
  Kline~Struhl, Mich\`{e}le~B. Nuijten, Julia~M. Rohrer, Felipe Romero, Anne~M.
  Scheel, Laura~D. Scherer, Felix~D. Sch\"{o}nbrodt, and Simine Vazire.
\newblock Replicability, robustness, and reproducibility in psychological
  science.
\newblock {\em Annual Review of Psychology}, 73(1):719--748, 2022.
\newblock PMID: 34665669.

\bibitem{pineau2021improving}
Joelle Pineau, Philippe Vincent-Lamarre, Koustuv Sinha, Vincent Larivi{\`e}re,
  Alina Beygelzimer, Florence d’Alch{\'e} Buc, Emily Fox, and Hugo
  Larochelle.
\newblock Improving reproducibility in machine learning research.
\newblock {\em Journal of Machine Learning Research}, 22, 2021.

\bibitem{gundersen2018state}
Odd~Erik Gundersen and Sigbj{\o}rn Kjensmo.
\newblock State of the art: Reproducibility in artificial intelligence.
\newblock In {\em Proceedings of the AAAI Conference on Artificial
  Intelligence}, volume~32, 2018.

\bibitem{salsabil2022scik}
Lamia Salsabil, Jian Wu, Muntabir~Hasan Choudhury, William~A. Ingram, Edward~A.
  Fox, Sarah~M. Rajtmajer, and C.~Lee Giles.
\newblock A study of computational reproducibility using urls linking to open
  access datasets and software.
\newblock In {\em Companion Proceedings of the Web Conference 2022}, WWW '22,
  page 784–788, New York, NY, USA, 2022. Association for Computing Machinery.

\bibitem{pimentel2019jupyter}
João~Felipe Pimentel, Leonardo Murta, Vanessa Braganholo, and Juliana Freire.
\newblock A large-scale study about quality and reproducibility of jupyter
  notebooks.
\newblock In {\em 2019 IEEE/ACM 16th International Conference on Mining
  Software Repositories (MSR)}, pages 507--517, 2019.

\bibitem{cascadetabnet}
Devashish Prasad, Ayan Gadpal, Kshitij Kapadni, Manish Visave, and Kavita
  Sultanpure.
\newblock Cascadetabnet: An approach for end to end table detection and
  structure recognition from image-based documents.
\newblock In {\em Proceedings of the IEEE/CVF conference on computer vision and
  pattern recognition workshops}, pages 572--573, 2020.

\bibitem{gatos2005automatic}
Basilios Gatos, Dimitrios Danatsas, Ioannis Pratikakis, and Stavros~J
  Perantonis.
\newblock Automatic table detection in document images.
\newblock In {\em International Conference on Pattern Recognition and Image
  Analysis}, pages 609--618. Springer, 2005.

\bibitem{DeepDsrt}
Sebastian Schreiber, Stefan Agne, Ivo Wolf, Andreas Dengel, and Sheraz Ahmed.
\newblock Deepdesrt: Deep learning for detection and structure recognition of
  tables in document images.
\newblock In {\em 2017 14th IAPR International Conference on Document Analysis
  and Recognition (ICDAR)}, volume~01, pages 1162--1167, 2017.

\bibitem{Multi-Type-TD-TSR}
Pascal Fischer, Alen Smajic, Giuseppe Abrami, and Alexander Mehler.
\newblock Multi-type-td-tsr--extracting tables from document images using a
  multi-stage pipeline for table detection and table structure recognition:
  From ocr to structured table representations.
\newblock In {\em German Conference on Artificial Intelligence (K{\"u}nstliche
  Intelligenz)}, pages 95--108. Springer, 2021.

\bibitem{Graph-based-TSR-lee}
Eunji Lee, Jaewoo Park, Hyung~Il Koo, and Nam~Ik Cho.
\newblock Deep-learning and graph-based approach to table structure
  recognition.
\newblock {\em Multimedia Tools and Applications}, 81(4):5827--5848, 2022.

\bibitem{collberg2016repeatability}
Christian Collberg and Todd~A. Proebsting.
\newblock Repeatability in computer systems research.
\newblock {\em Commun. ACM}, 59(3):62–69, feb 2016.

\bibitem{liu2021se}
Chao Liu, Cuiyun Gao, Xin Xia, David Lo, John~C. Grundy, and Xiaohu Yang.
\newblock On the reproducibility and replicability of deep learning in software
  engineering.
\newblock {\em {ACM} Trans. Softw. Eng. Methodol.}, 31(1):15:1--15:46, 2022.

\bibitem{jupyter-reproducibility}
Jo{\~a}o~Felipe Pimentel, Leonardo Murta, Vanessa Braganholo, and Juliana
  Freire.
\newblock A large-scale study about quality and reproducibility of jupyter
  notebooks.
\newblock In {\em 2019 IEEE/ACM 16th International Conference on Mining
  Software Repositories (MSR)}, pages 507--517. IEEE, 2019.

\bibitem{kamphuis2020bm25}
Chris Kamphuis, Arjen P~de Vries, Leonid Boytsov, and Jimmy Lin.
\newblock Which bm25 do you mean? a large-scale reproducibility study of
  scoring variants.
\newblock In {\em European Conference on Information Retrieval}, pages 28--34.
  Springer, 2020.

\bibitem{seibold2021computational}
Heidi Seibold, Severin Czerny, Siona Decke, Roman Dieterle, Thomas Eder,
  Steffen Fohr, Nico Hahn, Rabea Hartmann, Christoph Heindl, Philipp Kopper,
  Dario Lepke, Verena Loidl, Maximilian Mandl, Sarah Musiol, Jessica Peter,
  Alexander Piehler, Elio Rojas, Stefanie Schmid, Hannah Schmidt, Melissa
  Schmoll, Lennart Schneider, Xiao-Yin To, Viet Tran, Antje V{\"o}lker, Moritz
  Wagner, Joshua Wagner, Maria Waize, Hannah Wecker, Rui Yang, Simone Zellner,
  and Malte Nalenz.
\newblock A computational reproducibility study of plos one articles featuring
  longitudinal data analyses.
\newblock {\em PLOS ONE}, 16(6):1--15, 06 2021.

\bibitem{Prenkaj}
Bardh Prenkaj, Paola Velardi, Damiano Distante, and Stefano Faralli.
\newblock A reproducibility study of deep and surface machine learning methods
  for human-related trajectory prediction.
\newblock In {\em Proceedings of the 29th ACM International Conference on
  Information \& Knowledge Management}, CIKM '20, page 2169–2172, New York,
  NY, USA, 2020. Association for Computing Machinery.

\bibitem{stevens2020recommendations}
Laura~M. Stevens, Bobak~J. Mortazavi, Rahul~C. Deo, Lesley Curtis, and David~P.
  Kao.
\newblock Recommendations for reporting machine learning analyses in clinical
  research.
\newblock {\em Circulation: Cardiovascular Quality and Outcomes},
  13(10):e006556, 2020.

\bibitem{musgrave2020metric}
Kevin Musgrave, Serge Belongie, and Ser-Nam Lim.
\newblock A metric learning reality check.
\newblock In Andrea Vedaldi, Horst Bischof, Thomas Brox, and Jan-Michael Frahm,
  editors, {\em Computer Vision -- ECCV 2020}, pages 681--699, Cham, 2020.
  Springer International Publishing.

\bibitem{gobel2013icdartable}
Max G\"{o}bel, Tamir Hassan, Ermelinda Oro, and Giorgio Orsi.
\newblock Icdar 2013 table competition.
\newblock In {\em 2013 12th International Conference on Document Analysis and
  Recognition}, pages 1449--1453, 2013.

\bibitem{gao2019icdartable}
Liangcai Gao, Yilun Huang, Hervé Déjean, Jean-Luc Meunier, Qinqin Yan,
  Yu~Fang, Florian Kleber, and Eva Lang.
\newblock Icdar 2019 competition on table detection and recognition (ctdar).
\newblock In {\em 2019 International Conference on Document Analysis and
  Recognition (ICDAR)}, pages 1510--1515, 2019.

\bibitem{deeptabstr}
Shoaib~Ahmed Siddiqui, Imran~Ali Fateh, Syed Tahseen~Raza Rizvi, Andreas
  Dengel, and Sheraz Ahmed.
\newblock Deeptabstr: deep learning based table structure recognition.
\newblock In {\em 2019 International Conference on Document Analysis and
  Recognition (ICDAR)}, pages 1403--1409. IEEE, 2019.

\bibitem{ReS2TIM}
Wenyuan Xue, Qingyong Li, and Dacheng Tao.
\newblock Res2tim: Reconstruct syntactic structures from table images.
\newblock In {\em 2019 International Conference on Document Analysis and
  Recognition (ICDAR)}, pages 749--755, 2019.

\bibitem{qasim2019rethinking}
Shah~Rukh Qasim, Hassan Mahmood, and Faisal Shafait.
\newblock Rethinking table recognition using graph neural networks.
\newblock In {\em 2019 International Conference on Document Analysis and
  Recognition (ICDAR)}, pages 142--147. IEEE, 2019.

\bibitem{splerge}
Chris Tensmeyer, Vlad~I. Morariu, Brian Price, Scott Cohen, and Tony Martinez.
\newblock Deep splitting and merging for table structure decomposition.
\newblock In {\em 2019 International Conference on Document Analysis and
  Recognition (ICDAR)}, pages 114--121, 2019.

\bibitem{guided}
Khurram~Azeem Hashmi, Didier Stricker, Marcus Liwicki, Muhammad~Noman Afzal,
  and Muhammad~Zeshan Afzal.
\newblock Guided table structure recognition through anchor optimization.
\newblock {\em IEEE Access}, 9:113521--113534, 2021.

\bibitem{GRU-TSR}
Saqib~Ali Khan, Syed Muhammad~Daniyal Khalid, Muhammad~Ali Shahzad, and Faisal
  Shafait.
\newblock Table structure extraction with bi-directional gated recurrent unit
  networks.
\newblock In {\em 2019 International Conference on Document Analysis and
  Recognition (ICDAR)}, pages 1366--1371. IEEE, 2019.

\bibitem{TabStructNet}
Sachin Raja, Ajoy Mondal, and CV~Jawahar.
\newblock Table structure recognition using top-down and bottom-up cues.
\newblock In {\em European Conference on Computer Vision}, pages 70--86.
  Springer, 2020.

\bibitem{TGRNet}
Wenyuan Xue, Baosheng Yu, Wen Wang, Dacheng Tao, and Qingyong Li.
\newblock Tgrnet: A table graph reconstruction network for table structure
  recognition.
\newblock In {\em Proceedings of the IEEE/CVF International Conference on
  Computer Vision}, pages 1295--1304, 2021.

\bibitem{qiao2021lgpma}
Liang Qiao, Zaisheng Li, Zhanzhan Cheng, Peng Zhang, Shiliang Pu, Yi~Niu, Wenqi
  Ren, Wenming Tan, and Fei Wu.
\newblock Lgpma: Complicated table structure recognition with local and global
  pyramid mask alignment.
\newblock In {\em International Conference on Document Analysis and
  Recognition}, pages 99--114. Springer, 2021.

\bibitem{GTE}
Xinyi Zheng, Douglas Burdick, Lucian Popa, Xu~Zhong, and Nancy Xin~Ru Wang.
\newblock Global table extractor (gte): A framework for joint table
  identification and cell structure recognition using visual context.
\newblock In {\em Proceedings of the IEEE/CVF winter conference on applications
  of computer vision}, pages 697--706, 2021.

\bibitem{DBLP:conf/esann/JainPSV21}
Arushi Jain, Shubham Paliwal, Monika Sharma, and Lovekesh Vig.
\newblock {TSR-DSAW:} table structure recognition via deep spatial association
  of words.
\newblock In {\em 29th European Symposium on Artificial Neural Networks,
  Computational Intelligence and Machine Learning, {ESANN} 2021, Online event
  (Bruges, Belgium), October 6-8, 2021}, 2021.

\bibitem{li2021rethinking}
Yibo Li, Yilun Huang, Ziyi Zhu, Lemeng Pan, Yongshuai Huang, Lin Du, Zhi Tang,
  and Liangcai Gao.
\newblock Rethinking table structure recognition using sequence labeling
  methods.
\newblock In {\em International Conference on Document Analysis and
  Recognition}, pages 541--553. Springer, 2021.

\bibitem{dutta2019via}
Abhishek Dutta and Andrew Zisserman.
\newblock The via annotation software for images, audio and video.
\newblock In {\em Proceedings of the 27th ACM international conference on
  multimedia}, pages 2276--2279, 2019.

\bibitem{mchugh2012kappa}
M.~L. McHugh.
\newblock Interrater reliability: the kappa statistic.
\newblock {\em Biochem Med (Zagreb)}, 22(3):276--82, 2012.

\end{thebibliography}

\end{document}